\documentclass[11pt]{article}

\usepackage[T1]{fontenc}
\usepackage{lmodern}
\usepackage[margin=1in]{geometry}
\usepackage[round,authoryear]{natbib}

\newcommand{\model}{p_0}
\newcommand{\adapted}{p_\theta}
\newcommand{\controlled}{q_\theta}
\newcommand{\targetset}{\mathcal{A}}
\newcommand{\vocab}{\mathcal{V}}
\newcommand{\wmrule}{\mathcal{R}}
\newcommand{\indicator}{\mathbf{1}}
\newcommand{\nativeMass}{M_t}
\newcommand{\residualBias}{\alpha_t}
\newcommand{\watermarkability}{\mathsf{W}}

\usepackage{amsmath,amssymb,mathtools}
\usepackage{booktabs}
\usepackage{graphicx}
\usepackage{microtype}
\usepackage{multirow}
\usepackage{float}
\usepackage[section]{placeins}
\usepackage{xcolor}
\usepackage{enumitem}
\usepackage{tikz}
\usetikzlibrary{arrows.meta,positioning}
\usepackage[colorlinks,linkcolor=orange, anchorcolor=blue, citecolor=blue]{hyperref}
\usepackage{url}

\definecolor{wmblue}{HTML}{167D8D}
\definecolor{wmlight}{HTML}{E8F4F3}
\definecolor{wmorange}{HTML}{C65D32}

\title{Watermarkability: Measuring and Improving Language-Model Compatibility with Logit-Based Watermarks}

\author{
Zewei Deng\textsuperscript{1} \quad
Mohamed Seif\textsuperscript{2,3} \quad
Liyan Xie\textsuperscript{1}\\[0.55em]
\small \textsuperscript{1}Department of Industrial and Systems Engineering, University of Minnesota\\
\small \textsuperscript{2}Department of Computer Science and Engineering, Oakland University\\
\small \textsuperscript{3}Princeton University
}

\date{}

\begin{document}
\maketitle

\begin{abstract}
Language-model watermarks can help identify model-generated text, encode provenance information, and detect post-generation edits. A broad class of logit-based watermarks increases the logits of a designated token set at each decoding step. Because a base model's native probability mass on this set varies across contexts, models, and watermark rules, the same logit bias can yield different adherence rates and increases in perplexity (PPL). We introduce \emph{watermarkability} to characterize the compatibility between a model and the token sets requested by a watermark. At a target mass $\tau$, we quantify this property through a \emph{watermarkability cost}: the expected minimum KL divergence from the base-model distribution required to assign probability at least $\tau$ to the requested set. A lower cost means that the model is more watermarkable for that request distribution. To reduce this cost, we propose a two-stage method. In the first stage, we train a target-set-conditioned low-rank adapter on unions drawn from a 16-way vocabulary partition to increase the requested set's probability mass. In the second stage, a closed-form controller applies the smallest remaining uniform logit bias needed to reach the target mass. We evaluate the method on six watermark schedules using OPT-125M and Qwen3-8B. Across the 12 model--schedule pairs, the pre-adaptation watermarkability cost correlates with the log-PPL increase under a near-hard fixed-bias baseline (pooled Pearson $r=0.929$). For every schedule, at least 99.80\% of sampled tokens fall in the target set requested at their decoding step. At this reliability, the adapter and controller reduce average per-schedule prompt-conditional PPL by 46.3\% on OPT-125M and 5.2\% on Qwen3-8B relative to the fixed-bias baseline.
\let\thefootnote\relax\footnotetext{Code is available at
\url{https://github.com/zeweid1221/LLM_Watermarkability}.}
\end{abstract}

\section{Introduction}
\label{sec:introduction}

Many logit-based watermarks for language models modify generation by adding a bias to the logits of a selected token set. Specifically, let $\vocab$ denote the model vocabulary and $x_{1:T}\in\vocab^T$ denote a generated token sequence, and write $x_{<t}=(x_1,\ldots,x_{t-1})$ for the prefix at decoding step $t$. A watermark rule selects a target set $\targetset_t\subseteq\vocab$ and adds a scalar bias $\delta$ to every token in $\targetset_t$. If the base distribution mass of the next token $x_t$ is $\model(v\mid x_{<t})$ for $v\in \vocab$, the resulting watermarked distribution is
\begin{equation}\label{eq:watermark}
q_{\delta}(v\mid x_{<t})
=
\frac{\model(v\mid x_{<t})\exp\{\delta\indicator[v\in\targetset_t]\}}
{\sum_{u\in\vocab}\model(u\mid x_{<t})\exp\{\delta\indicator[u\in\targetset_t]\}}.    
\end{equation}
This form covers green/red-list watermarks \citep{kirchenbauer2023watermark,zhaoprovable}, multi-bit watermarks \citep{yoo2024mpac,xu2026xmark}, and code-constrained or pattern-based watermarks \citep{chao2024watermarking,xie2025combinatorial,deng2026detecting,deng2026anchorecc}. The effect of $\delta$ depends on the probability that the base model already assigns to $\targetset_t$. Consequently, the same bias can produce different adherence rates and perplexity changes across contexts, models, and watermark rules. 
This motivates our central research question: what is the minimum distributional change needed to reach a prescribed target-set probability, and can it be reduced while keeping the requested sets fixed?

\begin{figure}[t]
\centering
\includegraphics[width=\linewidth]{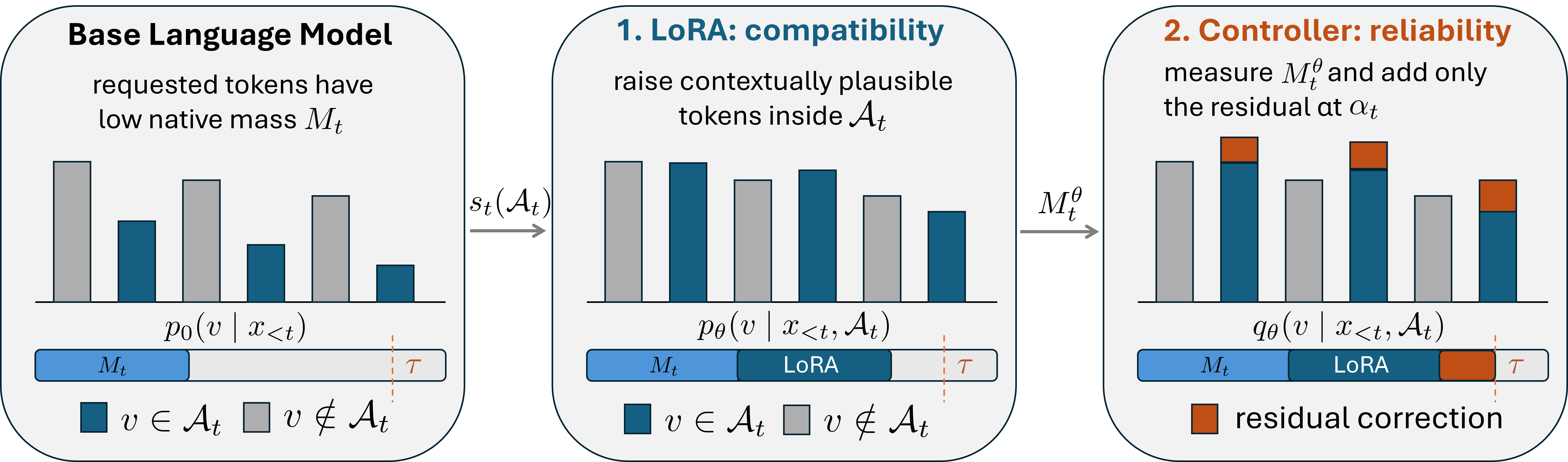}
\caption{Two-stage watermark realization. The target-conditioned LoRA redistributes probability toward contextually plausible tokens inside the requested set. The controller then applies a common residual logit shift, preserving the adapted token scores while bringing aggregate target mass from $M_t^\theta$ to $\tau$. Aligned illustrative gauges compare native, adapted, and controlled target mass.}
\label{fig:overview}
\end{figure}

To answer this question, we introduce \emph{watermarkability} as the compatibility between the base language model and a watermarking rule. For a given context and requested set $\targetset_t$, we define the watermarkability cost at target mass $\tau$ as the minimum KL divergence from the base-model distribution over all next-token distributions that assign probability at least $\tau$ to $\targetset_t$. We average this cost over the contexts and target sets induced by the watermark rule. A lower cost means that the model is more watermarkable: it can reach the prescribed target mass with less distributional distortion. The native probability mass is the total probability that the unmodified model assigns to all tokens in $\targetset_t$ before watermarking. When this mass is small, enforcing the request tends to select lower-probability tokens and thereby increase PPL under the base model.
To reduce this cost without modifying the watermark rule, we propose a two-stage method (see Figure~\ref{fig:overview}). To make one adapter reusable across schedules, we partition each model's vocabulary into 16 disjoint subsets and express each evaluated target set as a union of these subsets. Fixed schedules use prescribed unions, whereas prefix-dependent schedules select a new union from the prefix and key. The two stages serve complementary roles: the target-set-conditioned adapter raises the mass of contextually suitable tokens in $\targetset_t$, and the closed-form controller determines the smallest remaining nonnegative uniform logit bias $\alpha_t$ needed to reach target mass $\tau$.

We evaluate the proposed two-stage method on OPT-125M and Qwen3-8B using long-form question-answering prompts and six vocabulary-partition watermark schedules.
Our contributions are summarized as follows.
\enlargethispage{\baselineskip}
\begin{itemize}[leftmargin=*,itemsep=2pt,topsep=3pt]
    \item We characterize model--rule compatibility through watermarkability cost, the minimum distributional change needed to meet target-set requests at a prescribed reliability. Across 12 model--rule pairs, pairs with higher watermarkability cost generally exhibit larger fixed-bias log-PPL increases (Pearson $r=0.929$).

    \item We propose a two-stage method to reduce this cost. First, we train a target-set-conditioned low-rank adapter on generic unions from a 16-way vocabulary partition. Second, a closed-form controller applies the smallest remaining uniform logit bias needed to reach $\tau=0.99$. We evaluate the adapter on six held-out schedules spanning fixed, multi-symbol, and prefix-dependent designs.

\item Across all six schedules, the two-stage method maintains at least 99.80\% sampled-token adherence on both models. Relative to the near-hard $\delta=20$ baseline, it reduces mean per-schedule prompt-conditional PPL by 46.3\% on OPT-125M and 5.2\% on Qwen3-8B. PPL decreases for all six OPT schedules and four of six Qwen schedules.
\end{itemize}

\section{Related Work}

\paragraph{Vocabulary-partition and structured watermarking.}
Green-list watermarks use a key and usually the generated prefix to select a preferred vocabulary subset, increase its logits, and detect an excess of preferred tokens \citep{kirchenbauer2023watermark,kirchenbauer2023reliability}. Fixed-list, entropy-gated, semantically balanced, and context-adaptive variants modify how that subset is formed or activated \citep{zhaoprovable,lee2024sweet,liu2024adaptive,guo2024context}, while multi-bit methods select among several color classes to encode payload symbols \citep{yoo2024mpac,xu2026xmark}. Other constructions impose local patterns or code constraints, including error-correcting-code, combinatorial, and synchronization-assisted VT watermarks \citep{chao2024watermarking,xie2025combinatorial,deng2026detecting,deng2026anchorecc}. Across these schedules, our framework adapts the model to each step's preferred token set while leaving the request sequence and detector to the watermark rule.

\paragraph{Adaptive intervention and probability-aware partitioning.}
The detectability--quality trade-off of logit-based watermarking is well established \citep{pang2024no}, and recent methods show that a fixed bias need not represent the same effective strength across settings. Adaptive-bias approaches calibrate intervention using a constrained KGW objective, native green-list mass, or sequence-level logit heterogeneity \citep{cai2026statistical,wang2025morphmark,moon2026dynamicbias}. textGrain constructs an unbiased watermark by coupling keyed randomness to vocabulary blocks while constraining the resulting loss of sampling entropy \citep{li2026textgrain}. A complementary line improves the requested set itself: SSG constructs logit-balanced green and red lists \citep{gu2026ssg}, WaterMod partitions probability ranks into balanced classes \citep{park2026watermod}, and semantic or context-aware methods retain plausible alternatives within each partition \citep{liu2024adaptive,guo2024context}. These methods calibrate the intervention or redesign the partition within a particular watermark construction. We instead measure compatibility for externally specified target-set requests, train one adapter across set compositions, and use closed-form control to meet target-mass requirements.

\paragraph{Learning watermarked generation.}
Watermark distillation trains a model to reproduce a particular watermarked teacher \citep{gu2024learnability}. Our adapter instead conditions at runtime on the current requested set, supporting different keys, symbols, set cardinalities, and prefix-dependent schedules over shared base buckets. We implement this conditioning with low-rank adaptation \citep{hu2022lora} and an input-side summary of the requested set, combining cross-schedule training with an analytic reliability controller.

\section{Watermarkability}
\label{sec:formalization}

\paragraph{A single-step view of watermarkability.}
Consider a decoding step $t$ with prefix $x_{<t}$, where the watermark rule requests a target set $\targetset_t$ and requires the model to assign it probability at least $\tau$. How readily the model can satisfy this request depends on how much probability it already assigns to $\targetset_t$. We use \emph{watermarkability} to describe this compatibility: at a fixed $\tau$, a request that requires less adjustment to the model's next-token distribution has higher watermarkability.

We denote the base model's native probability mass on $\targetset_t$ by $\nativeMass
=
\sum_{v\in\targetset_t}
\model(v\mid x_{<t})$. 
After applying a uniform bias $\delta$, let
$M_t(\delta):=q_\delta(\targetset_t\mid x_{<t})$
denote the resulting target mass, so that $M_t(0)=M_t$. Eq.~\eqref{eq:watermark} gives the minimum nonnegative bias required to reach target mass $\tau$:
\begin{equation}
\delta_t^{(\tau)}
=
\left[
\operatorname{logit}(\tau)
-
\operatorname{logit}(\nativeMass)
\right]_{+},
\label{eq:min-delta}
\end{equation}
where $\operatorname{logit}(u)=\log(u/(1-u))$ and
$[a]_{+}=\max\{a,0\}$. Thus, $\delta_t^{(\tau)}$ measures the minimum logit adjustment needed to satisfy this request. However, the magnitude of this bias does not directly show how much the resulting next-token distribution differs from the base model. We therefore measure the cost of realizing the request by this distributional change: the smaller the minimum KL divergence needed to reach $\tau$, the higher the request's watermarkability.

\paragraph{The cost of realizing a request.}

When $\nativeMass\leq\tau$, Eq.~\eqref{eq:min-delta} gives
$M_t(\delta_t^{(\tau)})=\tau$, and the corresponding KL divergence is
\begin{equation}
\begin{aligned}
&D_{\mathrm{KL}}\Big(
q_{\delta_t^{(\tau)}}(\cdot\mid x_{<t})
\,\Vert\,
\model(\cdot\mid x_{<t})
\Big)=
\tau\delta_t^{(\tau)}
-\log\big(1-\nativeMass
+e^{\delta_t^{(\tau)}}\nativeMass\big)=
\tau\log\frac{\tau}{\nativeMass}
+(1-\tau)\log\frac{1-\tau}{1-\nativeMass}.
\end{aligned}
\label{eq:kl-cost}
\end{equation}
When $\nativeMass>\tau$, Eq.~\eqref{eq:min-delta} gives
$\delta_t^{(\tau)}=0$; the resulting distribution is then the base-model distribution itself, and the corresponding KL divergence is zero.

For any fixed value of
$q(\targetset_t\mid x_{<t})$, KL divergence from the base-model distribution is minimized by preserving the base model's relative token probabilities within $\targetset_t$ and within its complement $\targetset_t^c$. Minimizing over
$q(\targetset_t\mid x_{<t})\geq\tau$ therefore yields the distribution produced by $\delta_t^{(\tau)}$. We define the minimum single-step cost as
\begin{equation}\label{eq:cost}
\begin{aligned}
c_t(\tau):=
\min_{q:\,q(\targetset_t\mid x_{<t})\geq\tau}
D_{\mathrm{KL}}
\big(
q(\cdot\mid x_{<t})
\,\Vert\,
\model(\cdot\mid x_{<t})
\big)=
D_{\mathrm{KL}}
\Big(
q_{\delta_t^{(\tau)}}(\cdot\mid x_{<t})
\,\Vert\,
\model(\cdot\mid x_{<t})
\Big).
\end{aligned}
\end{equation}
Thus, $c_t(\tau)=0$ when $\nativeMass\geq\tau$ and is given by Eq.~\eqref{eq:kl-cost} otherwise. Notably, the cost depends on the specific target set: two sets of the same size can have different costs if the base model assigns them different native probability masses.

\paragraph{Watermarkability of a model--rule pair.}
For a fixed base model $p_0$ and watermark rule $\wmrule$, let
$\mathcal{D}_{\model,\wmrule}$ denote the joint distribution of prefixes and requested target sets under the specified context-sampling protocol and the rule's key and payload randomness. We define the watermarkability cost at target mass $\tau$ as
\begin{equation}
\watermarkability_{\model,\wmrule}(\tau)
=
\mathbb{E}_{(x_{<t},\targetset_t)\sim
\mathcal{D}_{\model,\wmrule}}
\left[c_t(\tau)\right].
\label{eq:watermarkability}
\end{equation}
At a fixed $\tau$, a lower value means that the model can realize the rule's requests with less distributional change and is therefore more watermarkable for that rule. Unlike a comparison at a fixed bias $\delta$, this definition fixes the target mass and allows the required bias $\delta_t^{(\tau)}$ to vary across contexts and requests. When a requested set has little native probability mass, enforcing it tends to select tokens assigned lower probability by the base model and can therefore increase perplexity. Watermarkability cost provides a pre-generation measure of model--rule compatibility at a matched target mass, while PPL measures the quality of the continuation ultimately produced under watermarking.

\section{A Two-Stage Method for Reducing Watermarking Cost}
\label{sec:method}

The definition in Section~\ref{sec:formalization} quantifies the cost of realizing a target-set request. We now describe a two-stage method for reducing this cost while preserving the watermark's schedule and detectability.
At each decoding step, the watermark rule specifies a target set $\targetset_t$. In the first stage, a low-rank adapter conditioned on $\targetset_t$ is trained to increase the probability assigned to this set. In the second stage, a closed-form controller applies the smallest remaining uniform logit bias needed to reach target mass $\tau$. Neither stage changes the target set given by the watermark rule. The adapter is intended to reduce the residual bias required by the controller, and we evaluate its effect on prompt-conditional perplexity empirically. Figure~\ref{fig:overview} summarizes the method.

\subsection{Vocabulary Partition and Training Requests}

We partition each model's vocabulary into $L=16$ disjoint \emph{base buckets} $\mathcal{B}_1,\ldots,\mathcal{B}_{16}$. Each base bucket is one of the smallest disjoint vocabulary subsets used by our construction. A watermark's target set is formed by taking a union of these buckets:
\begin{equation*}
    \targetset_t=\bigcup_{\ell\in J_t}\mathcal{B}_\ell,
    \qquad J_t\subseteq\{1,\ldots,L\}.
\end{equation*}
The index set $J_t$ may be fixed, follow a payload or code schedule, or be recomputed from the prefix and key. This gives several alphabet sizes and target rates without retraining a separate adapter for every materialized partition. A finer partition is also needed for the one-eighth boundary allocation in Anchor-ECC \citep{deng2026anchorecc}: we assign approximately one eighth of the usable vocabulary to the boundary set and divide the remainder equally between two payload sets. Here the choice of $L=16$ represents the smallest power-of-two granularity that realizes this allocation with whole base buckets, using $2$ buckets for boundaries and $7$ for each payload symbol, while also providing many keyed unions for the KGW variants.

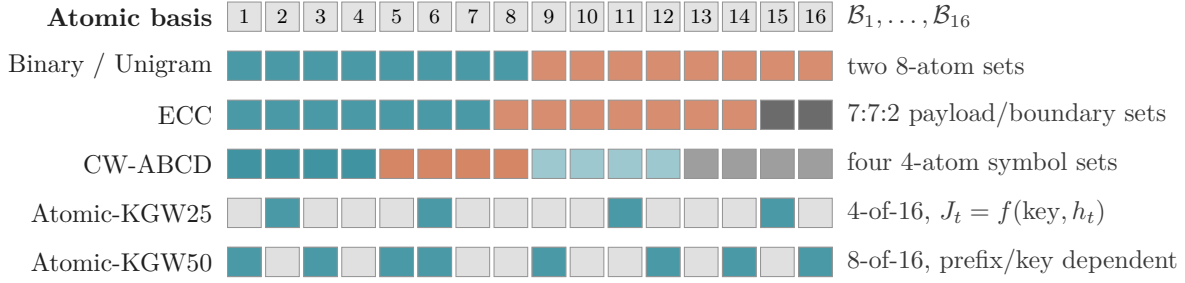
\begin{figure}[t]
\centering
\resizebox{0.96\linewidth}{!}{\begin{tikzpicture}[
  font=\small,
  atom/.style={draw=black!42,minimum width=0.46cm,minimum height=0.40cm,
               inner sep=0pt,font=\scriptsize},
  rowlabel/.style={anchor=east,font=\small,text=black!88},
  note/.style={anchor=west,font=\small,text=black!78}
]

\node[rowlabel,font=\small\bfseries] at (-0.30,2.85) {Atomic basis};
\foreach \i in {1,...,16} {
  \pgfmathsetmacro{\x}{0.52*(\i-1)}
  \node[atom,fill=black!11] at (\x,2.85) {\i};
}
\node[note] at (8.10,2.85) {$\mathcal B_1,\ldots,\mathcal B_{16}$};

\node[rowlabel] at (-0.30,2.18) {Binary / Unigram};
\foreach \i in {1,...,16} {
  \pgfmathsetmacro{\x}{0.52*(\i-1)}
  \ifnum\i<9
    \node[atom,fill=wmblue!78] at (\x,2.18) {};
  \else
    \node[atom,fill=wmorange!70] at (\x,2.18) {};
  \fi
}
\node[note] at (8.10,2.18) {two $8$-atom sets};

\node[rowlabel] at (-0.30,1.51) {ECC};
\foreach \i in {1,...,16} {
  \pgfmathsetmacro{\x}{0.52*(\i-1)}
  \ifnum\i<8
    \node[atom,fill=wmblue!78] at (\x,1.51) {};
  \else\ifnum\i<15
    \node[atom,fill=wmorange!70] at (\x,1.51) {};
  \else
    \node[atom,fill=black!58] at (\x,1.51) {};
  \fi\fi
}
\node[note] at (8.10,1.51) {$7{:}7{:}2$ payload/boundary sets};

\node[rowlabel] at (-0.30,0.84) {CW-ABCD};
\foreach \i in {1,...,16} {
  \pgfmathsetmacro{\x}{0.52*(\i-1)}
  \ifnum\i<5
    \node[atom,fill=wmblue!82] at (\x,0.84) {};
  \else\ifnum\i<9
    \node[atom,fill=wmorange!75] at (\x,0.84) {};
  \else\ifnum\i<13
    \node[atom,fill=wmblue!42] at (\x,0.84) {};
  \else
    \node[atom,fill=black!38] at (\x,0.84) {};
  \fi\fi\fi
}
\node[note] at (8.10,0.84) {four $4$-atom symbol sets};

\node[rowlabel] at (-0.30,0.17) {Atomic-KGW25};
\foreach \i in {1,...,16} {
  \pgfmathsetmacro{\x}{0.52*(\i-1)}
  \node[atom,fill=black!12] at (\x,0.17) {};
}
\foreach \i in {2,6,11,15} {
  \pgfmathsetmacro{\x}{0.52*(\i-1)}
  \node[atom,fill=wmblue!78] at (\x,0.17) {};
}
\node[note] at (8.10,0.17) {$4$-of-$16$, $J_t=f(\mathrm{key},h_t)$};

\node[rowlabel] at (-0.30,-0.50) {Atomic-KGW50};
\foreach \i in {1,...,16} {
  \pgfmathsetmacro{\x}{0.52*(\i-1)}
  \node[atom,fill=black!12] at (\x,-0.50) {};
}
\foreach \i in {1,3,5,6,9,12,14,16} {
  \pgfmathsetmacro{\x}{0.52*(\i-1)}
  \node[atom,fill=wmblue!78] at (\x,-0.50) {};
}
\node[note] at (8.10,-0.50) {$8$-of-$16$, prefix/key dependent};

\end{tikzpicture}}
\caption{One model-specific 16-bucket partition supports fixed and dynamic target-set compositions. Structured schedules regroup base buckets into symbol sets; dynamic schedules recompute $J_t$ from the prefix and key. Highlighted buckets show one generation step.}
\label{fig:atomic-compositions}
\end{figure}

Table~\ref{tab:native-schedules} summarizes the native target-set split of each evaluated watermark schedule before its realization with the shared 16-bucket partition. Figure~\ref{fig:atomic-compositions} then shows how these native splits are assembled from the 16 base buckets.

\begin{table}[H]
\centering
\caption{Native bucket structures of the evaluated watermarks, grouped by the number of bucket classes. Vocabulary shares follow the bucket-label order.}
\label{tab:native-schedules}
\small
\setlength{\tabcolsep}{4.5pt}
\renewcommand{\arraystretch}{1.12}
\begin{tabular*}{0.98\linewidth}{@{\extracolsep{\fill}}lcll@{}}
\toprule
Watermarks & \multicolumn{3}{c}{Native bucket partition} \\
\cmidrule(lr){2-4}
& Classes & Bucket labels & Vocabulary share \\
\midrule
Alternating binary & \multirow{3}{*}{2} & Symbol 0 / Symbol 1 & $1/2$, $1/2$ \\
Unigram, Grouped-KGW50 & & Target / Complement & $1/2$, $1/2$ \\
Grouped-KGW25 & & Target / Complement & $1/4$, $3/4$ \\
\specialrule{0.35pt}{2pt}{2pt}
CW-ABCD & 4 & A / B / C / D & $1/4$ each \\
\specialrule{0.35pt}{2pt}{2pt}
Anchor-ECC (quality variant) & 3 & Payload 0 / 1 / Boundary anchor & $7/16$, $7/16$, $1/8$ \\
\bottomrule
\end{tabular*}
\end{table}

The base buckets are designed to be model-specific in our experiments. Specifically, for a given base model $p_0$, we collect next-token probabilities on $N$ calibration contexts and optimize soft assignments $a_{v\ell}$ from every usable token $v\in\mathcal V$ to bucket $\ell$. For context $i$, let $m_i(\ell)$ be the probability mass assigned to bucket $\ell$. At grouping resolution $g\in\{2,4,8,16\}$, $m_i^{(g)}(j)=\sum_{\ell=(j-1)L/g+1}^{jL/g}m_i(\ell)$ denotes the mass of the $j$-th consecutive group of $L/g$ base buckets, for $j=1,\ldots,g$. The objective balances grouped masses while regularizing token counts, semantic coverage, and assignment entropy:
\begin{equation}
\mathcal{L}_{\mathrm{part}}=
\sum_{g\in\{2,4,8,16\}}\frac{1}{Ng}\sum_{i=1}^{N}\sum_{j=1}^{g}
\left(m_i^{(g)}(j)-\frac{1}{g}\right)^2
+\lambda_c\mathcal{L}_{\mathrm{count}}
+\lambda_s\mathcal{L}_{\mathrm{sem}}
+\lambda_e\mathcal{L}_{\mathrm{ent}}.
\label{eq:partition-loss}
\end{equation}
Here $\mathcal L_{\mathrm{count}}$ balances token counts across buckets, $\mathcal L_{\mathrm{sem}}$ distributes each semantic and native-probability group across buckets, and $\mathcal L_{\mathrm{ent}}$ encourages discrete assignments. Appendix~\ref{app:partition-details} gives their exact definitions. We materialize the optimized assignment into a token lookup table. Figure~\ref{fig:atomic-compositions} shows how representative fixed and dynamic schedules compose target sets from these base buckets.

\subsection{Target-Set Conditioning and LoRA}

Using the partition above, we train a shared adapter on target sets formed by unions of $2$, $4$, $7$, or $8$ base buckets. For each cardinality, the training curriculum is organized into complete passes over all possible bucket unions. We implement the adapter with low-rank adaptation (LoRA), which augments a frozen linear map $W$ with a trainable rank-$r$ residual, mapping $Wh$ to $Wh+(\gamma/r)UVh$ for learned low-rank factors $U$ and $V$ and scale $\gamma$ \citep{hu2022lora}. We insert these updates into the query and value projections of every self-attention layer. The query updates affect how the current position weights information from the prefix, while the value updates affect the content carried through attention. All original language-model parameters remain frozen, and the trainable parameters consist of the query/value LoRA factors and a target-set projector described below. Both components use rank $r=16$, with LoRA scale $\gamma=32$.

At decoding step $t$, the watermark rule specifies the current target set $\targetset_t$. We summarize how this set intersects the base model's plausible next-token candidates before passing the request to the shared adapter. Let $\mathcal T_t^{(k)}$ contain the $k$ tokens with largest probabilities under $\model(\cdot\mid x_{<t})$, and let $e_v$ denote the frozen input embedding of token $v$. Within this set, $\widetilde p_t(v)=\model(v\mid x_{<t})/\sum_{u\in\mathcal T_t^{(k)}}\model(u\mid x_{<t})$ is the renormalized base-model probability, and $\bar e_v=e_v/\lVert e_v\rVert_2$ is the normalized input embedding. We use $k=128$ and form
\begin{equation*}
s_t=\operatorname{norm}\!\left(
\sum_{v\in\mathcal T_t^{(k)}}\widetilde p_t(v)
\bigl(2\cdot \indicator[v\in\targetset_t]-1\bigr)\bar e_v
\right).
\end{equation*}
Here $\operatorname{norm}(u)=u/\lVert u\rVert_2$. The indicator assigns positive signs to requested candidates and negative signs to the remaining candidates, and the probability weighting gives more influence to plausible next tokens. The resulting vector summarizes the semantic contrast between likely tokens admitted by the current request and likely tokens outside it.

A rank-$r$ condition projector maps this summary to the model hidden dimension as $c_t=U_c\operatorname{SiLU}(D_c\operatorname{LayerNorm}(s_t))$ and adds $c_t$ to the input representation used to predict the next token. The conditioned representation then passes through the LoRA-equipped query and value projections, producing the adapted next-token distribution $\adapted(v\mid x_{<t},\targetset_t)$, where $\theta$ comprises the condition-projector and LoRA parameters. The parameters $\theta$ are shared across decoding steps and schedules; the conditioning vector is recomputed from the current target set. The controller uses the exact membership mask of $\targetset_t$ to calibrate target mass in the next stage.

\subsection{Closed-Form Reliability Controller}
\label{sec:controller}

After adaptation, the target mass is
\begin{equation*}
M_t^\theta=\sum_{v\in\targetset_t}\adapted(v\mid x_{<t},\targetset_t),
\end{equation*}
and we denote $z_t^\theta(v)$ as the corresponding adapted-model logit for token $v$. Adding a common shift $\alpha$ to all target-set logits changes this mass to
\begin{equation}
M_t^\theta(\alpha)=
\frac{e^\alpha M_t^\theta}
{1-M_t^\theta+e^\alpha M_t^\theta}.
\label{eq:shifted-mass}
\end{equation}
Solving $M_t^\theta(\alpha)=\tau$ and applying no positive shift when the adapted mass already exceeds $\tau$ gives
\begin{equation}
\residualBias=
\left[\operatorname{logit}(\tau)-\operatorname{logit}(M_t^\theta)\right]_+,
\qquad
z_t^{\mathrm{final}}(v)=z_t^\theta(v)+\residualBias\indicator[v\in\targetset_t].
\label{eq:controller}
\end{equation}
Let $\controlled$ denote the normalized next-token distribution induced by $z_t^{\mathrm{final}}$. Because Eq.~\eqref{eq:shifted-mass} is monotone in $\alpha$, Eq.~\eqref{eq:controller} is the smallest nonnegative common shift that yields target mass at least $\tau$ on $\targetset_t$.
Applying the same shift to every requested token leaves their pairwise logit differences unchanged, so the controller preserves the adapter's ranking within $\targetset_t$. The adapter determines which requested tokens fit the context; the controller chooses the remaining aggregate mass needed for reliability.

\subsection{Training Objective}

We generate constrained trajectories from the generic composition curriculum and update only the adapter parameters $\theta$. Each
training position $t$ provides a target set $\targetset_t$ and a realized token $y_t\in\targetset_t$. We minimize the per-position
loss
\begin{equation*}
\mathcal{L}_t=
-\log \controlled(y_t\mid x_{<t},\targetset_t)
+\lambda_r\residualBias^2
+\lambda_{\mathrm{KL}}D_{\mathrm{KL}}
\left(\adapted(\cdot\mid x_{<t},\targetset_t)\,\Vert\,\model(\cdot\mid x_{<t})\right),
\end{equation*}
where $\lambda_r,\lambda_{\mathrm{KL}}\ge 0$ are weighting
coefficients. The first term is the negative log-likelihood (NLL)
of the realized token under the controlled distribution and
encourages natural choices within the target set. The second term penalizes the squared residual bias $\residualBias^2$, which
encourages the adapter itself to place probability mass on the
target set before control is applied. The KL term limits drift of the adapter from the frozen model. Writing $\mathcal I$ for the set of active training positions, we optimize the average per-position loss, $\mathcal L_{\mathrm{train}}(\theta)=|\mathcal I|^{-1}\sum_{t\in\mathcal I}\mathcal L_t$, using the corresponding tokenwise average in each minibatch.
\section{Experiments}
\label{sec:experiments}

\subsection{Experimental Setup}

\paragraph{Models and data.}
We evaluate OPT-125M \citep{zhang2022opt} and Qwen3-8B \citep{yang2025qwen3}. Prompts are English long-form questions derived from ELI5/LFQA \citep{fan2019eli5}. For each model, base-bucket partition construction uses 5,000 calibration contexts and 500 validation contexts. Adapter preparation pairs every prompt once with each of the four target cardinalities, producing 20,000 generic composition trajectories and 2,000 validation trajectories. Evaluation uses 256 held-out prompts per watermark schedule.

\paragraph{Schedules.}
The six primary vocabulary-partition watermark schedules are: a three-symbol ECC schedule with seven payload positions and one boundary; an alternating binary schedule; the four-symbol ACADBCBD combinatorial pattern \citep{xie2025combinatorial}; a fixed balanced Unigram-style green list \citep{zhaoprovable}; and prefix-dependent 8-of-16 and 4-of-16 grouped KGW. The ECC evaluation uses the Anchor-ECC schedule \citep{deng2026anchorecc}; synchronization-string VT watermarking \citep{deng2026detecting} induces a closely related code-constrained bucket pattern. The grouped KGW rules retain prefix- and key-dependent target selection while selecting complete base buckets; we use the name Grouped-KGW to distinguish them from native token-level KGW. The complete schedule registry and auxiliary results appear in Appendix~\ref{app:rules}.

\paragraph{Training and generation.}
Both experiments use six training epochs, $\tau=0.99$, 96 generated tokens, temperature 0.75, and full-softmax sampling. This target requires nearly all probability mass to lie in the requested set while retaining a $1\%$ tolerance for tokens outside it. Setting $\tau=1$ would require eliminating all probability outside the requested set and hence an infinitely large logit shift whenever the model assigns any mass outside it. The OPT adapter has 614,400 trainable parameters; the Qwen adapter has 7,798,784. We report corpus-level prompt-conditional PPL under each frozen base model and compare it only within that model.

\paragraph{Operating points.}
\textsc{Fixed-$\delta$} adds bias 20 to the target set, following the approximate-hard configuration in Anchor-ECC \citep{deng2026anchorecc}. Across the twelve primary model--schedule pairs, it achieves 99.94--100\% target-set adherence (Appendix~\ref{app:complete-results}), providing a high-reliability comparison with the controller target $\tau=0.99$. \textsc{Controller} applies Eq.~\eqref{eq:controller} to the frozen model. \textsc{LoRA-only} uses learned adaptation without residual control. \textsc{LoRA+Controller} combines both. Token adherence is the fraction of active positions whose sampled token belongs to $\targetset_t$. Exact-block rate requires every request in a structured block to be realized.

\subsection{Watermarkability and Generation Cost}

We first test whether pre-adaptation watermarkability cost predicts the generation cost of different watermark requests. For each model, we generate one unwatermarked continuation for each of the 256 held-out LFQA prompts and evaluate all six schedules on the same teacher-forced next-token distributions. Thus, the text and model distributions remain fixed while the requested target set changes. We denote the finite-sample average of Eq.~\eqref{eq:watermarkability} by $\widehat{\mathsf W}(\tau)$. Table~\ref{tab:cross-rule-wability} compares $\widehat{\mathsf W}(0.99)$ with the conditional log-PPL increase measured from the corresponding fixed-$\delta=20$ generations.

\begin{table}[ht!]
\centering
\caption{Pre-adaptation watermarkability cost and observed fixed-bias generation cost on matched prompts. $\Delta\log\mathrm{PPL}=\log(\mathrm{PPL}_{\delta=20}/\mathrm{PPL}_{\mathrm{clean}})$; negative values indicate lower fixed-bias PPL.}
\label{tab:cross-rule-wability}
\small
\setlength{\tabcolsep}{5.0pt}
\begin{tabular}{lrrrr}
\toprule
& \multicolumn{2}{c}{OPT-125M} & \multicolumn{2}{c}{Qwen3-8B} \\
\cmidrule(lr){2-3}\cmidrule(lr){4-5}
Rule & $\widehat{\mathsf W}(0.99)$ & $\Delta\log\mathrm{PPL}$
     & $\widehat{\mathsf W}(0.99)$ & $\Delta\log\mathrm{PPL}$ \\
\midrule
ECC              & 1.354 &  1.140 & 3.303 & 2.159 \\
Binary           & 0.957 &  0.341 & 2.373 & 1.397 \\
CW-ABCD          & 1.956 &  1.120 & 4.604 & 2.370 \\
Unigram          & 1.144 & $-0.090$ & 2.343 & 1.274 \\
Grouped-KGW50    & 0.943 &  0.637 & 2.459 & 1.587 \\
Grouped-KGW25    & 1.934 &  1.304 & 4.597 & 2.578 \\
\bottomrule
\end{tabular}
\end{table}

Across the six rules, $\widehat{\mathsf W}(0.99)$ tracks observed log-PPL increase on Qwen (Pearson $r=0.955$, Spearman $\rho=0.943$) and OPT ($r=0.741$, $\rho=0.600$); pooling the twelve model--rule points gives $r=0.929$ and $\rho=0.930$. For every rule, Qwen has both higher watermarkability cost and a larger PPL increase relative to its own clean baseline. Fixed-bias PPL ranges from $3.58\times$ to $13.18\times$ the clean value on Qwen, compared with $0.91\times$ to $3.68\times$ on OPT. The two 4-of-16 schedules are the most expensive Qwen requests. This agreement shows why the same nominal bias does not define a comparable operating point across different base models.

\subsection{Reliability-Matched Composition-Pool Results}

Table~\ref{tab:cross-model} reports the reliability-matched comparison. We train one adapter on generic base-bucket unions and evaluate it on six held-out watermark schedules. LoRA+Controller reaches at least 99.80\% adherence for every primary rule on both models. All six OPT schedules improve (46.3\% mean per-rule PPL reduction), as do four Qwen schedules (5.2\% mean reduction).

\begin{table}[ht!]
\centering
\caption{Held-out results after generic composition-pool training. $\Delta$ is the relative PPL reduction from fixed $\delta=20$ to LoRA+Controller; positive is better. The Mean-row $\Delta$ averages the six per-rule relative reductions.
}
\label{tab:cross-model}
\small
\setlength{\tabcolsep}{5pt}
\begin{tabular}{lrrrrrr}
\toprule
& \multicolumn{3}{c}{OPT-125M} & \multicolumn{3}{c}{Qwen3-8B}\\
\cmidrule(lr){2-4}\cmidrule(lr){5-7}
Rule & Fixed & LoRA+Ctrl. & $\Delta$ & Fixed & LoRA+Ctrl. & $\Delta$ \\
\midrule
ECC              & 26.57 & 20.73 & 22.0\% & 17.92 & 20.19 & $-12.7$\% \\
Binary           & 11.95 & 4.37  & 63.4\% & 8.37  & 7.52  & 10.1\% \\
CW-ABCD          & 26.03 & 13.00 & 50.1\% & 22.15 & 20.38 & 8.0\% \\
Unigram          & 7.76  & 3.22  & 58.6\% & 7.40  & 5.43  & 26.6\% \\
Grouped-KGW50    & 16.06 & 7.50  & 53.3\% & 10.11 & 9.69  & 4.2\% \\
Grouped-KGW25    & 31.29 & 21.79 & 30.4\% & 27.26 & 28.69 & $-5.2$\% \\
\midrule
Mean             & 19.94 & 11.77 & 46.3\% & 15.53 & 15.32 & 5.2\% \\
\cmidrule(lr){1-7}
Minimum adherence & \multicolumn{3}{c}{99.80\%} & \multicolumn{3}{c}{99.80\%} \\
\bottomrule
\end{tabular}
\end{table}

Paired bootstrap over the 256 prompts gives 95\% intervals of 43.6--48.6\% on OPT and 0.6--9.3\% on Qwen for the mean per-rule reduction. On Qwen, Binary and Unigram improve with intervals above zero, ECC increases PPL, and the intervals for CW-ABCD and both Grouped-KGW settings include zero. The Qwen results therefore vary more across target-set schedules than the OPT results. At the same operating point, the exact-block rates for ECC, CW, and Binary are 98.44\%, 98.44\%, and 99.65\% on OPT and 98.63\%, 98.60\%, and 99.63\% on Qwen.

Table~\ref{tab:components} separates adaptation from residual control. LoRA alone lowers PPL but reaches 46.4\% macro adherence on OPT and 44.3\% on Qwen. Adding the controller raises adherence to 99.83\% and 99.82\%, while reducing residual bias from 4.78 to 3.64 and from 5.09 to 4.11, respectively.

\begin{table}[H]
\centering
\caption{Component-level macro results; $\overline{\alpha}$ is the mean residual bias over active decoding steps.}
\label{tab:components}
\small
\begin{tabular}{llrrr}
\toprule
Model & Method & PPL & Adherence (\%) & Mean residual bias $\overline{\alpha}$ \\
\midrule
\multirow{3}{*}{OPT-125M}
& Controller & 19.92 & 99.78 & 4.78 \\
& LoRA only & 5.66 & 46.43 & --- \\
& LoRA + controller & 11.77 & 99.83 & 3.64 \\
\midrule
\multirow{3}{*}{Qwen3-8B}
& Controller & 15.57 & 99.76 & 5.09 \\
& LoRA only & 5.49 & 44.34 & --- \\
& LoRA + controller & 15.32 & 99.82 & 4.11 \\
\bottomrule
\end{tabular}
\end{table}

\section{Discussion and Conclusion}

This work studied how compatible a language model is with the token sets requested by logit-based watermarks. We measured this compatibility by the watermarkability cost, the expected minimum KL
divergence from the base model needed to place probability at least $\tau$ on each requested set. To reduce this cost without changing the requested sets, we trained a target-set-conditioned LoRA adapter on unions of sixteen vocabulary buckets and combined it with a closed-form controller that adds only the remaining logit bias needed to reach $\tau$. On OPT-125M and Qwen3-8B with six watermark schedules, the pre-adaptation cost tracked the PPL increase of
fixed-bias watermarking (pooled Pearson $r=0.929$), and every schedule was more costly for Qwen3-8B than for OPT-125M. With adherence at or above 99.80\%, our method reduced PPL relative to the fixed bias by 46.3\% on OPT-125M and by 5.2\% on Qwen3-8B.

Future work includes evaluating more models, prompt domains, and quality measures, and improving the gains on larger models through richer set encodings or training focused on the schedules a deployed watermark actually uses. It would also be useful to extend watermarkability to watermarks that do not act through a target set, and to use the cost before deployment to choose a watermark rule and target mass for a given model.

\bibliography{references}
\bibliographystyle{plainnat}

\appendix
\section{Target Schedules and Composition Coverage}
\label{app:rules}

\begin{table}[H]
\centering
\caption{Watermark schedules used only for held-out evaluation. ``Dynamic'' means the exact base-bucket union changes with the generated prefix.}
\small
\begin{tabular}{llll}
\toprule
Rule & Target composition & Schedule & Dynamic \\
\midrule
ECC & 7:7:2 buckets & feasible 7-bit payload + boundary & No \\
Binary & 8:8 buckets & alternating symbols & No \\
CW-ABCD & four 4-bucket sets & ACADBCBD & No \\
Unigram & fixed 8-of-16 & constant green set & No \\
Grouped-KGW50 & 8-of-16 & prefix/key-seeded union & Yes \\
Grouped-KGW25 & 4-of-16 & prefix/key-seeded union & Yes \\
Entropy-KGW & 8-of-16 & entropy-gated prefix/key union & Yes \\
\bottomrule
\end{tabular}
\end{table}

Adapter training uses the generic cardinality curriculum described in Section~\ref{sec:method}; none of the schedule labels above enters the training trajectories. Table~\ref{tab:coverage} reports its exact target-set coverage.

\begin{table}[H]
\centering
\caption{Distinct base-bucket unions covered by the generic composition curriculum.}
\label{tab:coverage}
\small
\begin{tabular}{lrrr}
\toprule
Target size & Observed & Possible & Coverage \\
\midrule
2-of-16 & 120 & 120 & 100.0\% \\
4-of-16 & 1,820 & 1,820 & 100.0\% \\
7-of-16 & 11,440 & 11,440 & 100.0\% \\
8-of-16 & 12,870 & 12,870 & 100.0\% \\
\bottomrule
\end{tabular}
\end{table}

For ECC, a 96-token continuation contains twelve blocks, each comprising seven payload requests and one boundary request. Its payload schedule is drawn from the intersection of a length-7 VT constraint and a Hamming parity constraint. CW repeats its eight-symbol pattern twelve times. Exact-block rate requires every position in a block to realize its requested set.

\section{Complete Composition-Pool Results}
\label{app:complete-results}

\begin{table}[H]
\centering
\caption{Complete OPT-125M evaluation after generic composition-pool training. Adherence is shown as a fraction.}
\vspace{4pt}
\small
\begin{tabular*}{0.92\textwidth}{@{\extracolsep{\fill}}lrrrrrrrr@{}}
\toprule
& \multicolumn{2}{c}{Fixed $\delta=20$} & \multicolumn{2}{c}{Controller} & \multicolumn{2}{c}{LoRA only} & \multicolumn{2}{c}{LoRA+Controller}\\
\cmidrule(lr){2-3}\cmidrule(lr){4-5}\cmidrule(lr){6-7}\cmidrule(lr){8-9}
Rule & PPL & Adh. & PPL & Adh. & PPL & Adh. & PPL & Adh. \\
\midrule
ECC          & 26.57 & 1.0000 & 26.38 & 0.9972 & 5.91 & 0.4675 & 20.73 & 0.9980 \\
Binary       & 11.95 & 1.0000 & 11.87 & 0.9979 & 4.99 & 0.5952 & 4.37  & 0.9983 \\
CW-ABCD      & 26.03 & 1.0000 & 26.03 & 0.9974 & 6.84 & 0.3183 & 13.00 & 0.9980 \\
Unigram      & 7.76  & 1.0000 & 8.16  & 0.9980 & 5.52 & 0.4969 & 3.22  & 0.9989 \\
Grouped-KGW50 & 16.06 & 1.0000 & 15.94 & 0.9980 & 4.68 & 0.5935 & 7.50  & 0.9982 \\
Grouped-KGW25 & 31.29 & 1.0000 & 31.14 & 0.9984 & 6.03 & 0.3141 & 21.79 & 0.9983 \\
Entropy-KGW  & 11.18 & 1.0000 & 11.16 & 0.9977 & 5.05 & 0.6082 & 6.18  & 0.9979 \\
\bottomrule
\end{tabular*}
\end{table}

\begin{table}[H]
\centering
\caption{Complete Qwen3-8B evaluation after generic composition-pool training. Adherence is shown as a fraction.}
\vspace{4pt}
\small
\begin{tabular*}{0.92\textwidth}{@{\extracolsep{\fill}}lrrrrrrrr@{}}
\toprule
& \multicolumn{2}{c}{Fixed $\delta=20$} & \multicolumn{2}{c}{Controller} & \multicolumn{2}{c}{LoRA only} & \multicolumn{2}{c}{LoRA+Controller}\\
\cmidrule(lr){2-3}\cmidrule(lr){4-5}\cmidrule(lr){6-7}\cmidrule(lr){8-9}
Rule & PPL & Adh. & PPL & Adh. & PPL & Adh. & PPL & Adh. \\
\midrule
ECC          & 17.92 & 0.9994 & 17.80 & 0.9972 & 5.38 & 0.4539 & 20.19 & 0.9983 \\
Binary       & 8.37  & 0.9997 & 8.28  & 0.9975 & 4.61 & 0.5444 & 7.52  & 0.9982 \\
CW-ABCD      & 22.15 & 0.9998 & 22.91 & 0.9978 & 7.09 & 0.2901 & 20.38 & 0.9983 \\
Unigram      & 7.40  & 0.9999 & 7.47  & 0.9982 & 4.40 & 0.5280 & 5.43  & 0.9981 \\
Grouped-KGW50 & 10.11 & 0.9998 & 10.33 & 0.9977 & 4.62 & 0.5394 & 9.69  & 0.9980 \\
Grouped-KGW25 & 27.26 & 0.9997 & 26.59 & 0.9975 & 6.86 & 0.3048 & 28.69 & 0.9982 \\
Entropy-KGW  & 2.18  & 1.0000 & 2.17  & 0.9977 & 2.21 & 0.5242 & 2.40  & 0.9968 \\
\bottomrule
\end{tabular*}
\end{table}

\section{Illustrative Qwen Generations}
\label{app:qwen-examples}

To complement the aggregate PPL results, we present three matched Qwen cases
in which lower sequence-level NLL coincides with a visibly cleaner
continuation. Each comparison holds the prompt, decoding protocol, and target
reliability fixed; PPL is computed from the continuation's mean base-model
NLL. These selected cases illustrate the observed quality difference, while
the complete quantitative comparison is reported above. The excerpts are
taken directly from the saved outputs without language editing; ellipses
indicate omitted trailing text.

\paragraph{Binary, sequence 70.}
\textbf{Prompt (abridged).} What is the difference between a scientific theory and law?
Why is something repeatedly confirmed not ``upgraded'' into a law?

\noindent\textbf{Fixed $\delta=20$ (PPL 18.96; adherence 1.000).}
\begin{quote}\small
And if a law could be replaced with a new theory (like if a new theory
explains relatable things), how is that possible that the old law could still
work? Okay. So the user is asking about scientific laws, like $F=ma$ versus
theories like rel Okay. Let me try and work out this question. First, I
remember the user is asking about scientific laws, like $F=ma$ versus theories
like rel Okay. \ldots
\end{quote}

\noindent\textbf{LoRA+Controller (PPL 11.76; adherence 1.000).}
\begin{quote}\small
And if a law could be replaced with a law, like how Newton was superseded with
relavant laws, then why is that the case? Also for a law, like $F=ma$, it seems
like an equation that is a formula that is used for calculations. So if that's
not the case for a law, then what is a law? Because a law seems like a formula.
First, I'll start by addressing what I think you're confused by. \ldots
\end{quote}

\paragraph{Unigram, sequence 127.}
\textbf{Prompt (abridged).} Does the electromagnetic spectrum abruptly stop at gamma
rays, or are there still higher-energy, shorter-wavelength waves?

\noindent\textbf{Fixed $\delta=20$ (PPL 18.09; adherence 1.000).}
\begin{quote}\small
[Note: This was an actual physics exam from a few centuries ago, which was an
essay exam, so this was not a multiple-choice exam.] Okay, so my user asks a
few questions regarding gamma radiation, higher energy waves, their origins,
artificial production, etc. Let's dissect this. \ldots
\end{quote}

\noindent\textbf{LoRA+Controller (PPL 11.87; adherence 0.990).}
\begin{quote}\small
The electromagnetic spectrum does not abruptly stop at gamma-ray wavelengths,
even though gamma-ray wavelengths are considered to have higher energy.
The electromagnetic spectrum consists of electromagnetic waves. \ldots
\end{quote}

\paragraph{Grouped-KGW50, sequence 33.}
\textbf{Prompt (abridged).} If only one photon goes through a double slit, is there an
interference pattern on the other side, and why does observation change it?

\noindent\textbf{Fixed $\delta=20$ (PPL 14.93; adherence 1.000).}
\begin{quote}\small
Context: I've been trying to understand what exactly it is that the double
slit shows us in laymen terms, so I thought this novel might help me. I have so
far only seen this in the form that if you do this without observation the
pattern is interference pattern, but as soon the you observe it, the photons
go as individual and you don't observe the interference \ldots
\end{quote}

\noindent\textbf{LoRA+Controller (PPL 7.03; adherence 1.000).}
\begin{quote}\small
Okay, the idea that a single photon behaves simultaneously as a wave and a
particle and can have multiple probabilities of being in different states is
a concept from quantum mechanics. So if one photon is shot through a
two-slit apparatus, it's not like it splits into two particles and each
particle goes through one of the slits, but rather, the photon is in a
superposition of being in either \ldots
\end{quote}

\section{Schedule-Aware Training Ablation}
\label{app:schedule-aware-comparison}

For each model, we train one shared adapter on target sets generated by the seven evaluation schedule families. The adapter receives the current target set without a watermark identity or codeword label. This schedule-aware setting uses 35,000 training trajectories, compared with 20,000 generic composition trajectories in the main experiment. On OPT, all six primary schedules improve and the mean per-rule PPL reduction is 48.5\% (95\% paired-bootstrap interval: 46.1--50.6\%), compared with 46.3\% under generic composition training. On Qwen, the corresponding means are 8.2\% and 5.2\%; both settings improve four schedules in point estimate.

\begin{table}[H]
\centering
\caption{Complete corrected schedule-aware OPT-125M ablation. Adherence is shown as a fraction.}
\small
\begin{tabular*}{0.92\textwidth}{@{\extracolsep{\fill}}lrrrrrrrr@{}}
\toprule
& \multicolumn{2}{c}{Fixed $\delta=20$} & \multicolumn{2}{c}{Controller} & \multicolumn{2}{c}{LoRA only} & \multicolumn{2}{c}{LoRA+Controller}\\
\cmidrule(lr){2-3}\cmidrule(lr){4-5}\cmidrule(lr){6-7}\cmidrule(lr){8-9}
Rule & PPL & Adh. & PPL & Adh. & PPL & Adh. & PPL & Adh. \\
\midrule
ECC          & 26.57 & 1.0000 & 26.38 & 0.9972 & 7.55 & 0.5590 & 20.08 & 0.9977 \\
Binary       & 11.95 & 1.0000 & 11.87 & 0.9979 & 4.73 & 0.7694 & 4.33 & 0.9989 \\
CW-ABCD      & 26.03 & 1.0000 & 26.03 & 0.9974 & 8.26 & 0.3867 & 12.73 & 0.9985 \\
Unigram      & 7.76  & 1.0000 & 8.16  & 0.9980 & 4.05 & 0.7324 & 3.23  & 0.9991 \\
Grouped-KGW50 & 16.06 & 1.0000 & 15.94 & 0.9980 & 6.06 & 0.6509 & 7.82  & 0.9984 \\
Grouped-KGW25 & 31.29 & 1.0000 & 31.14 & 0.9984 & 7.00 & 0.3580 & 18.20 & 0.9985 \\
Entropy-KGW  & 11.18 & 1.0000 & 11.16 & 0.9977 & 5.79 & 0.6532 & 6.49  & 0.9979 \\
\bottomrule
\end{tabular*}
\end{table}

\paragraph{Schedule-aware OPT uncertainty.}
All six primary-rule intervals for PPL reduction lie above zero: ECC $[21.3,27.4]\%$, Binary $[59.7,67.3]\%$, CW-ABCD $[43.5,57.7]\%$, Unigram $[53.6,62.7]\%$, Grouped-KGW50 $[46.4,55.9]\%$, and Grouped-KGW25 $[33.7,49.2]\%$.

\begin{table}[H]
\centering
\caption{Complete schedule-aware Qwen3-8B ablation. Adherence is shown as a fraction.}
\small
\begin{tabular*}{0.92\textwidth}{@{\extracolsep{\fill}}lrrrrrrrr@{}}
\toprule
& \multicolumn{2}{c}{Fixed $\delta=20$} & \multicolumn{2}{c}{Controller} & \multicolumn{2}{c}{LoRA only} & \multicolumn{2}{c}{LoRA+Controller}\\
\cmidrule(lr){2-3}\cmidrule(lr){4-5}\cmidrule(lr){6-7}\cmidrule(lr){8-9}
Rule & PPL & Adh. & PPL & Adh. & PPL & Adh. & PPL & Adh. \\
\midrule
ECC          & 17.92 & 0.9994 & 17.80 & 0.9972 & 6.92 & 0.4681 & 21.38 & 0.9981 \\
Binary       & 8.37  & 0.9997 & 8.28  & 0.9975 & 5.47 & 0.5833 & 7.07  & 0.9984 \\
CW-ABCD      & 22.15 & 0.9998 & 22.91 & 0.9978 & 9.75 & 0.3160 & 19.63 & 0.9986 \\
Unigram      & 7.40  & 0.9999 & 7.47  & 0.9982 & 5.40 & 0.5929 & 4.59  & 0.9987 \\
Grouped-KGW50 & 10.11 & 0.9998 & 10.33 & 0.9977 & 5.66 & 0.5724 & 9.62  & 0.9983 \\
Grouped-KGW25 & 27.26 & 0.9997 & 26.59 & 0.9975 & 9.52 & 0.3217 & 27.56 & 0.9987 \\
Entropy-KGW  & 2.18  & 1.0000 & 2.17  & 0.9977 & 2.22 & 0.5506 & 2.36  & 0.9988 \\
\bottomrule
\end{tabular*}
\end{table}

\paragraph{Schedule-aware Qwen uncertainty.}
We bootstrap paired sequence NLLs with 20,000 resamples. Relative PPL changes from fixed $\delta=20$ to LoRA+Controller have 95\% intervals of $[+13.9,+25.0]\%$ for ECC, $[-23.1,-7.5]\%$ for Binary, $[-22.5,+1.5]\%$ for CW, $[-45.2,-29.6]\%$ for Unigram, $[-14.4,+5.3]\%$ for Grouped-KGW50, $[-10.1,+13.6]\%$ for Grouped-KGW25, and $[+2.2,+13.9]\%$ for Entropy-KGW. Positive values mean higher PPL.

\section{Controller Derivation}
\label{app:controller}

Let $M=p(\targetset)$ and add a common bias $\alpha$ to target-set logits. The normalizer is $Z=1-M+e^\alpha M$, and the resulting target mass is $e^\alpha M/Z$. Solving $e^\alpha M/Z=\tau$ gives
\begin{equation*}
    \alpha=\operatorname{logit}(\tau)-\operatorname{logit}(M).
\end{equation*}
When $M\geq\tau$, no positive intervention is needed, yielding the rectified expression in Eq.~\eqref{eq:controller}. Since the likelihood ratio is $e^\alpha/Z$ inside the set and $1/Z$ outside it,
\begin{align*}
D_{\mathrm{KL}}(q\Vert p)
&=q(\targetset)\log\frac{e^\alpha}{Z}
 +(1-q(\targetset))\log\frac{1}{Z}=\tau\alpha-\log Z.
\end{align*}

\section{Partition, Training, and Compute Details}
\label{app:partition-details}

For completeness, the regularizers used in Eq.~\eqref{eq:partition-loss} are
\begin{align*}
\mathcal L_{\mathrm{count}}
&=\frac{1}{L}\sum_{\ell=1}^{L}
\left(\frac{1}{|\mathcal V_u|}\sum_{v\in\mathcal V_u}a_{v\ell}-\frac{1}{L}\right)^2,\\
\mathcal L_{\mathrm{sem}}
&=\frac{1}{|\mathcal G|L}\sum_{G\in\mathcal G}\sum_{\ell=1}^{L}
\left(\frac{1}{|G|}\sum_{v\in G}a_{v\ell}-\frac{1}{L}\right)^2,\\
\mathcal L_{\mathrm{ent}}
&=-\frac{1}{|\mathcal V_u|}\sum_{v\in\mathcal V_u}\sum_{\ell=1}^{L}
a_{v\ell}\log a_{v\ell}.
\end{align*}
Here $\mathcal G$ groups tokens by the conjunction of an embedding-based semantic LSH signature and a native-probability stratum. The count term balances bucket sizes, the semantic term spreads every group across the $L$ buckets, and the entropy term makes the soft assignments discrete. These definitions match the partition optimizer used in our experiments.

Each model-specific base-bucket partition uses 5,000 LFQA calibration contexts and 500 validation contexts. The optimizer estimates the mass-balancing objective with minibatches of 256 calibration contexts. Before materializing its token-to-bucket lookup, it applies aggregation levels $K\in\{2,4,8,16\}$, eight-bit semantic LSH, and 256 native-mass strata. The partition-loss weights are $\lambda_c=0.25$, $\lambda_s=0.10$, and $\lambda_e=0.02$.

Both adapters use rank 16 (LoRA scale 32), dropout 0.05, learning rate $2\times10^{-4}$, weight decay 0.01, six epochs, and $\lambda_r=\lambda_{\mathrm{KL}}=0.02$; target summaries use the top 128 candidates. Evaluation covers seven rules, five modes, and 256 prompts per rule--mode pair (8,960 sequences per model). Including preparation, training, and evaluation, the composition-pool runs take 1.49 hours for OPT and approximately 8.6 hours for Qwen on one NVIDIA A100-SXM4-40GB GPU. Qwen selects its epoch-1 checkpoint by validation objective.

The auxiliary schedule-aware runs take 2.63 hours for OPT and 15.3 hours for Qwen on the same GPU. Qwen uses its epoch-1 checkpoint selected by validation objective.

\section{Broader Impact}

Lower-cost realization may aid provenance but also facilitate misuse. Deployment requires defenses against spoofing, key theft, removal, and false attribution, plus key management, wrong-key calibration, detector audits, and separation of provenance from content claims.

\end{document}